\documentclass[letterpaper, 10 pt, conference]{ieeeconf}  

\IEEEoverridecommandlockouts                              

\usepackage{float}
\usepackage{amssymb}
\usepackage{amsmath} 
\usepackage{graphicx}
\usepackage{booktabs}
\usepackage{array}
\usepackage{subcaption}
\let\labelindent\relax
\usepackage{enumitem}
\usepackage{caption}  
\usepackage{multirow} 
\usepackage{tabularx}
\usepackage{booktabs}
\usepackage{ragged2e}
\usepackage{threeparttable}
\usepackage{stfloats}
\usepackage{amsthm}    
\usepackage{hyperref}
\usepackage{cleveref}
\usepackage[top=0.75in, bottom=0.78in,left=0.75in,right=0.75in]{geometry}
\theoremstyle{definition}

\title{\LARGE \bf
Agile in the Face of Delay: Asynchronous End-to-End Learning for Real-World Aerial Navigation
}

\author{Yude Li$^{\dagger}$, Zhexuan Zhou$^{\dagger}$, Huizhe Li, Youmin Gong$^{*}$ and Jie Mei$^{*}$
\thanks{$\dagger$These authors contributed equally to this work.}
\thanks{*Corresponding authors: Jie Mei jmei@hit.edu.cn and Youmin Gong gongyoumin@hit.edu.cn.}
\thanks{The authors are with the School of Intelligence and Engineering, and also with the Guangdong Provincial Key Laboratory of Intelligent Morphing Mechanisms and Adaptive Robotics, Harbin
Institute of Technology, Shenzhen, Guangdong, China. This work was
supported in part by the Shenzhen Science and Technology Program under
Grant JCYJ20241202124010014, JCYJ20240813104923032, and GXWD20231129140908002.}
}

\begin{document}

\maketitle
\thispagestyle{empty}
\pagestyle{empty}

\begin{abstract}

Robust autonomous navigation for Autonomous Aerial Vehicles (AAVs) in complex environments is a critical capability. However, modern end-to-end navigation faces a key challenge: the high-frequency control loop needed for agile flight conflicts with low-frequency perception streams, which are limited by sensor update rates and significant computational cost. This mismatch forces conventional synchronous models into undesirably low control rates. To resolve this, we propose an asynchronous reinforcement learning framework that decouples perception and control, enabling a high-frequency policy to act on the latest IMU state for immediate reactivity, while incorporating perception features asynchronously. To manage the resulting data staleness, we introduce a theoretically-grounded Temporal Encoding Module (TEM) that explicitly conditions the policy on perception delays, a strategy complemented by a two-stage curriculum to ensure stable and efficient training. Validated in extensive simulations, our method was successfully deployed in zero-shot sim-to-real transfer on an onboard NUC, where it sustains a 100~Hz control rate and demonstrates robust, agile navigation in cluttered real-world environments. Our project details are available at \url{https://hitsz-mas.github.io/Agile-Asynch-Nav/}.
\end{abstract}


\section{INTRODUCTION}

Robust and agile autonomous navigation is crucial for the effective deployment of Autonomous Aerial Vehicles (AAVs) in complex and unstructured environments. Conventional approaches typically rely on a modular pipeline that separates the navigation task into distinct stages of perception~\cite{xu2022fast}, planning~\cite{zhou2022swarm}, and control~\cite{9917382}. While well-established, this architecture is prone to inter-module error propagation and significant decision-making latency, which can compromise flight safety and efficiency. In response to these challenges, end-to-end learning frameworks, particularly those leveraging reinforcement learning (RL)~\cite{kaufmann2023champion, loquercio2021learning}, have emerged as a promising alternative. By directly mapping sensory inputs to control actions, these methods can overcome the limitations of modular systems, thereby reducing latency and improving robustness.

\begin{figure}[!htb] 
    \centering
    \includegraphics[width=0.94\linewidth]{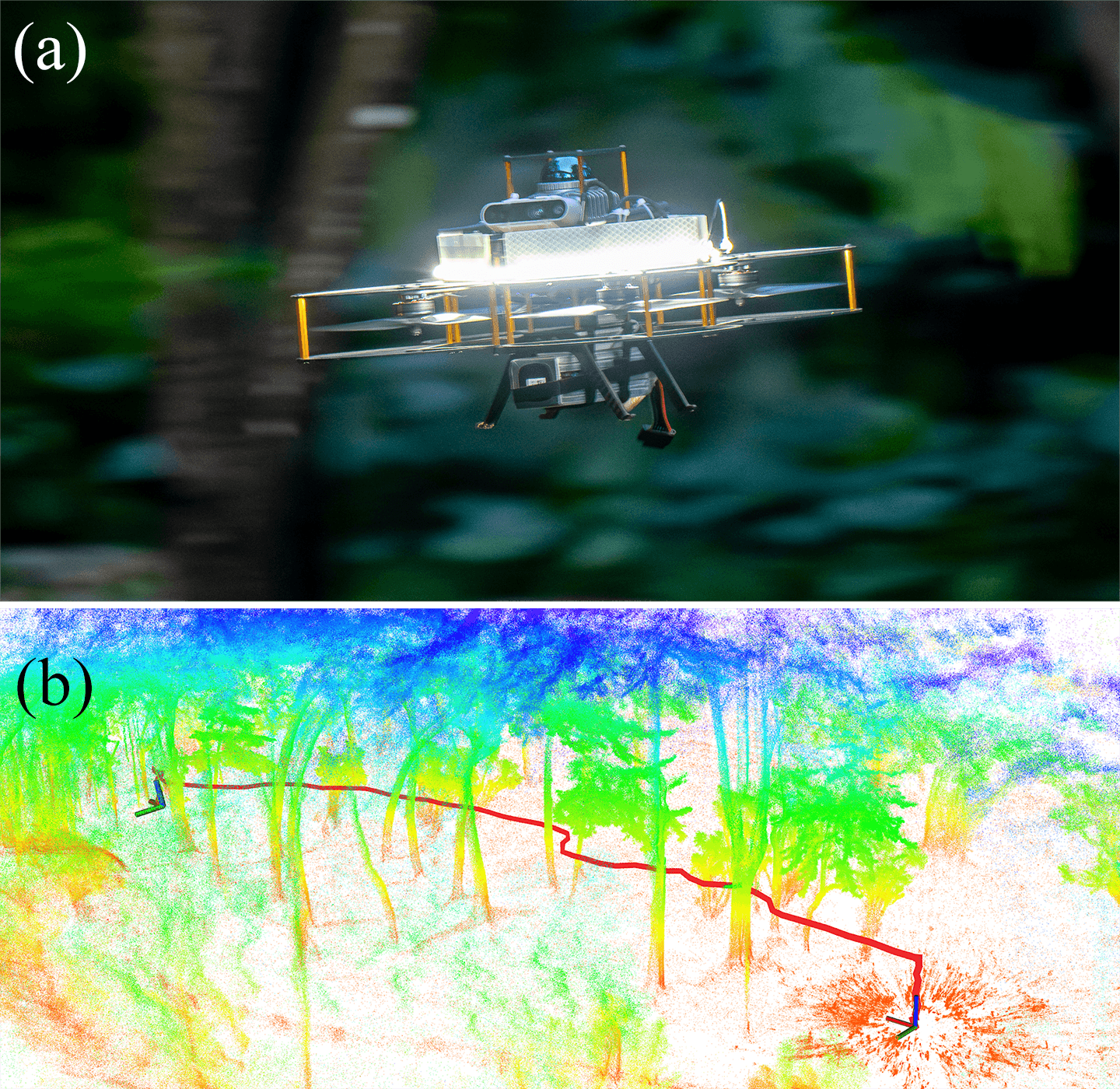} 
    
    \caption{The proposed asynchronous framework validated through zero-shot sim-to-real flight in a dense forest. (a) The custom-built AAV navigating through cluttered trees. (b) The complete, collision-free trajectory of a 30-meter flight test through the forest.}
    \label{fig:first}
    \vspace{-3mm} 
\end{figure}

However, the practical deployment of these end-to-end models on computationally constrained AAVs introduces a significant challenge. Unlike platforms with substantial onboard processing capabilities, such as legged robots~\cite{lee2020learning} or autonomous vehicles~\cite{chen2024end}, AAVs face a critical resource contention. They must arbitrate between processing high-bandwidth sensor data and maintaining the high-frequency control loops essential for agile flight. This contention manifests as a fundamental temporal mismatch: while high-frequency state information from an Inertial Measurement Unit (IMU) is readily available, the perception stream is simultaneously constrained by both the low native update rates of its sensors (e.g., LiDAR and cameras) and the heavy computational overhead of processing their data on resource-limited hardware. This asynchrony between perception and control has become a key barrier to achieving highly agile and robust end-to-end navigation.

This work presents a novel asynchronous framework designed to resolve the temporal mismatch between perception and control. The proposed method operates by decoupling these two loops, enabling the control policy to execute at a high frequency using the latest available state. This decoupling, in turn, introduces the consequential problem of data staleness, formally known as the Age of Information (AoI)~\cite{kaul2012real}, where the policy must rely on delayed perception features. This challenge is addressed through a principled Temporal Encoding Module (TEM), a mechanism that allows the network to explicitly reason about information delay and compensate for the resulting partial observability. The framework further incorporates an efficient pillar-based perception module to minimize latency at the sensor processing stage. Through extensive validation in the NVIDIA Isaac Sim environment and a successful zero-shot transfer to physical AAVs, this framework is shown to enable robust, high-frequency navigation (Fig.~\ref{fig:first}). The main contributions of this work are summarized as follows:
\begin{itemize}

\item A novel end-to-end network architecture is presented, featuring a computationally efficient LiDAR data processing module that leverages spatial features to enable agile flight in complex environments.
\item A theoretically-grounded Temporal Encoding Module is proposed to enable asynchronous policy inference. By modeling data staleness, this module resolves the partial observability induced by the low-frequency perception stream, facilitating robust and high-frequency control from low-frequency sensors on computationally limited platforms.
\item A two-stage curriculum learning strategy is applied to optimize the asynchronous policy, enabling successful zero-shot, sim-to-real transfer. This result, validated on physical AAVs, confirms the practical applicability and robustness of the proposed asynchronous framework.
\end{itemize}

\begin{figure*}[htb] 
\centering 
\includegraphics[width=0.92\textwidth]{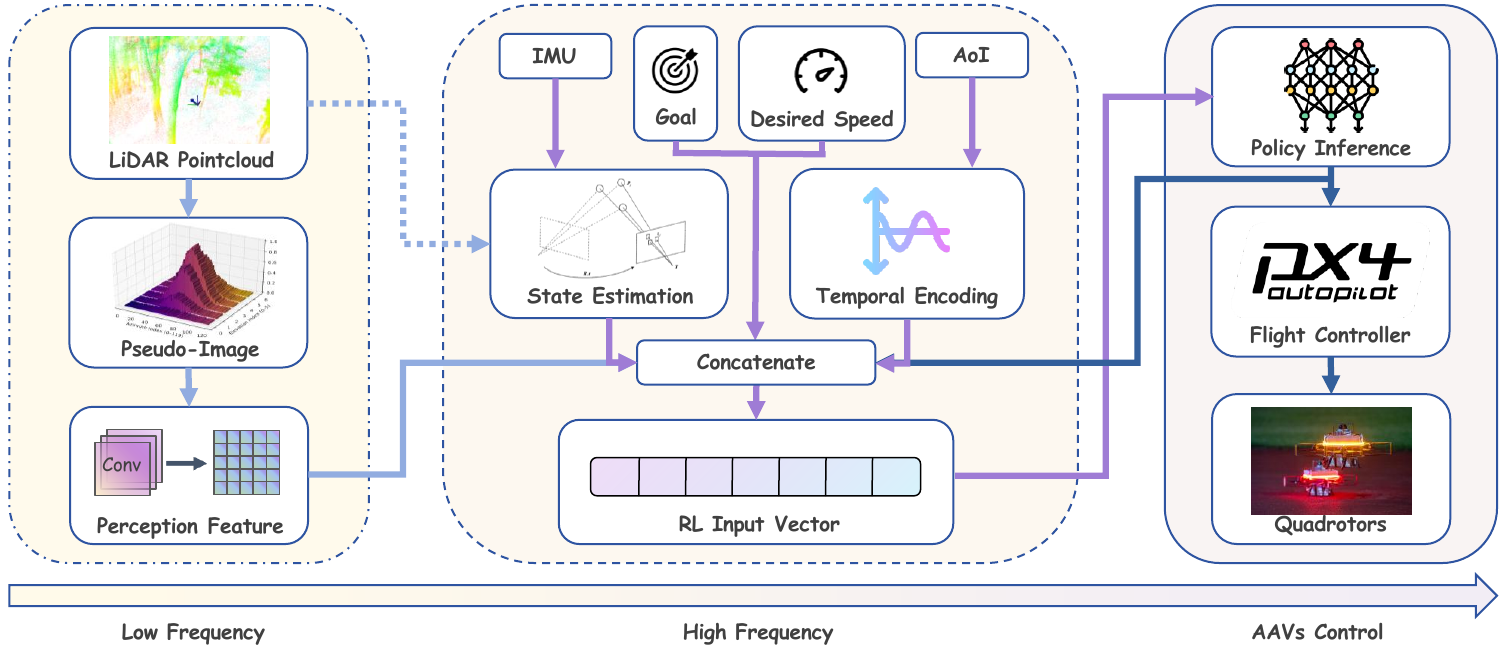} 
\caption{Overview of the asynchronous framework. The low-frequency perception module converts LiDAR point clouds to a pseudo-image, which a CNN processes into a feature vector. The high-frequency control module concatenates this feature with the latest IMU state, previous action, desired speed, and a temporal encoding vector representing data staleness. An MLP policy then processes this combined state to generate high-frequency control commands.}
\label{Fig.main2} 
\end{figure*}

\section{RELATED WORK}

Architectures for autonomous navigation have evolved from modular pipelines to learning-based paradigms to address error propagation and latency. In modular perception–planning–control pipelines, representative planners attain robust and agile flight yet remain exposed to inter-module error propagation and accumulated delays~\cite{ren2025safety, gao2020teach, liu2023integrated,11245959}.

Hybrid approaches integrate machine learning into specific system components, with examples in perception-aware imitation~\cite{tordesillas2023deep}, one-stage planning~\cite{lu2024you}, learned time allocation~\cite{wu2024deep}, and adaptive speed constraints~\cite{zhao2024learning}. Similarly, frameworks like NavRL employ a specialized perception module~\cite{xu2025navrl}. However, many of these hybrid architectures continue to use a synchronous update schedule. This model makes the decision-making latency directly dependent on the perception system's processing rate and the available onboard computing power.

End-to-end control via reinforcement learning demonstrates high-speed navigation~\cite{loquercio2021learning, zhang2025learning} and champion-level racing from visual inputs~\cite{kaufmann2023champion, xing2024bootstrapping}. In vision-based end-to-end methods, studies explore obstacle avoidance using learned motion cues or adaptive-speed control~\cite{song2023learning, hu2025seeing, yu2024mavrl}. An alternative line directly processes point clouds—spanning model-based~\cite{han2025neupan} and model-free methods~\cite{xu2025flying, fan2025flying}—and typically relies on point, voxel, or pillar encoders (PointNet++~\cite{qi2017pointnet++}, VoxelNet~\cite{zhou2018voxelnet}, PointPillars~\cite{8954311}) proven effective in autonomous driving and quadrupedal locomotion~\cite{lee2020learning,chen2024end}, suggesting transfer to AAVs. However, many end-to-end stacks typically operate under a single-inference-per-sensor-frame regime (policy updates gated by new perception frames, even if the low-level controller ticks faster), constraining control reactivity by sensor update rates and perception compute. Traditionally, sensor latency is addressed by decoupling perception and planning~\cite{gao2019flying} or employing high-frequency reactive policies~\cite{pantic2023obstacle}. Our framework adopts a similar decoupled structure within end-to-end learning, but introduces TEM to explicitly compensate for the resulting data staleness.

To address the challenges of asynchrony and delay in systems, multiple lines of research offer distinct approaches. For instance, while theories like Reinforcement Learning (RL) explicitly model perception-to-action delays~\cite{ramstedt2019real}, multi-rate and event-triggered control focuses on optimizing update schedules under resource constraints~\cite{heemels2012introduction}, and the concept of Age of Information (AoI) provides a metric to quantify data freshness~\cite{kaul2012real, yates2021age}. Collectively, these perspectives reveal that the temporal mismatch between perception and control is not merely an architectural issue of system design, but a fundamental structural problem rooted in physical constraints. This necessitates explicit temporal reasoning at the moment of decision-making. Yet, most work in this area has focused on low-level components like controllers and communication protocols. The integration of such temporal modeling into high-level, end-to-end perception-to-action policies remains a significant and underexplored challenge, which this work confronts.

Despite steady progress, most state-of-the-art end-to-end frameworks typically operate under a single-inference-per-sensor-frame schedule (e.g.,~\cite{song2023learning, hu2025seeing, yu2024mavrl, xu2025flying, fan2025flying}), effectively tying decision updates to perception throughput and overlooking data staleness, formally the Age of Information (AoI)~\cite{kaul2012real, yates2021age}, and the partial observability inherent in real asynchronous systems. We address this gap by decoupling perception and control and by explicitly encoding AoI at decision time via a Temporal Encoding Module, which enables robust high-frequency control under realistic low-rate sensing without relying solely on implicit temporal memory.

\section{METHODOLOGY}

Our proposed framework (Fig.~\ref{Fig.main2}) comprises a low-frequency perception pipeline and a high-frequency control loop that operate asynchronously. It transforms raw LiDAR data into a 2D Pseudo-Image (\ref{subsection:Pointcloud_Preprocessing_and_Representation}) and formulates the task as a Markov Decision Process (MDP) (\ref{subsection:Learning_to_Fly}). The core of our method is an Asynchronous End-to-End Network, whose temporal encoding is theoretically justified (\ref{subsection:Asynchronous end-to-end network}). The network's Architecture (\ref{subsection:Network_Architecture}) and Policy Training details (\ref{subsection:Policy_Training}) are also presented.

\subsection{Pseudo-Image Generation}
\label{subsection:Pointcloud_Preprocessing_and_Representation}

To process raw and unstructured point cloud data, a preliminary transformation into a structured pseudo-image representation is required. The proposed methodology employs a spherical coordinate projection for this conversion. This approach is specifically chosen for its efficacy in preserving range information, which is essential for subsequent collision avoidance tasks.

Each point in the raw point cloud, captured in the robot's body frame, can be represented in spherical coordinates as $\mathbf{p} = (r, \theta, \phi)$, where $r$ denotes the radial distance, $\theta$ the azimuthal angle in the XY-plane, and $\phi$ the polar angle measured from the positive Z-axis. The raw point cloud is the set of all points $\xi = \{\mathbf{p}_m \mid m = 1, \dots, N\}$, where each $\mathbf{p}_m = (r_m, \theta_m, \phi_m)$.

To convert the unstructured point cloud into a structured pseudo-image, the point cloud is first discretized into a 2D grid by partitioning the sensor's field of view into angular bins. Each pillar $P_{i,j}$ is then defined as the set of all points within a specific angular bin

\begin{equation}
    P_{i,j} = \{ \mathbf{p}\mid \mathbf{p} \in \xi, \theta_i \le \theta < \theta_{i+1}, \phi_j \le \phi < \phi_{j+1} \}.
\end{equation}
This grid spans the sensor's field of view, $[\theta_{\min}, \theta_{\max})$ and $[\phi_{\min}, \phi_{\max})$, with angular resolutions $\Delta\theta_{\text{pillar}}$ and $\Delta\phi_{\text{pillar}}$. This results in a grid of dimensions $(N_{\phi}, N_{\theta})$, where $N_{\theta} = \lfloor (\theta_{\max} - \theta_{\min}) / \Delta\theta_{\text{pillar}} \rfloor$ and $N_{\phi} = \lfloor (\phi_{\max} - \phi_{\min}) / \Delta\phi_{\text{pillar}} \rfloor$.

 The value of each pixel $I(i,j)$ in the pseudo-image is determined by the contents of its corresponding pillar $P_{i,j}$ according to the following function:
\begin{equation}
\label{eq:feature_extraction}
I(i, j) = 
\begin{cases} 
    \min \{r \mid \mathbf{p} = (r, \theta, \phi) \in P_{i,j}\} & \text{if } P_{i,j} \neq \varnothing \\
    r_{\text{max}} & \text{if } P_{i,j} = \varnothing 
\end{cases} ,
\end{equation}
where $r_{\text{max}}$ is the sensor's maximum range. The resulting collection of range values forms a single-channel pseudo-image of dimensions $(N_{\phi}, N_{\theta}, 1)$. This structured tensor serves as the input to a convolutional backbone for feature extraction.

\subsection{Reinforcement Learning Framework}
\label{subsection:Learning_to_Fly}


\subsubsection{Problem Formulation}
\label{subsection:Problem_Formulation}

 The navigation task is formulated as a Markov Decision Process (MDP), defined by the tuple $(\mathcal{S}, \mathcal{A}, P, R, \gamma)$, where $\mathcal{S}$ is the state space, $\mathcal{A}$ is the action space, $P$ is the state transition function, $R$ is the reward function, and $\gamma$ is the discount factor. The agent's objective is to learn a policy $\pi: \mathcal{S} \to \mathcal{A}$ that maximizes the expected cumulative discounted reward, given by

\begin{equation}
 G_t = \sum_{k=0}^{\infty} \gamma^k R_{t+k+1}.
\end{equation}

The state observation $s_t \in \mathcal{S}$ is a composite vector defined as the concatenation of perception features and the quadrotor's internal state:
\begin{equation}
\label{eq:state_vector}
s_t = [\mathbf{z}_t, \mathbf{p}_{\text{rel}}, \mathbf{q}, \mathbf{v}, \boldsymbol{\omega}, \mathbf{a}_{\text{prev}}, v_{\text{des}}, \Phi(\Delta t_{\text{lidar}})]^T,
\end{equation}
where $\mathbf{z}_t \in \mathbb{R}^{D_{\text{perception}}}$ is the perception feature,  $\mathbf{p}_{\text{rel}} \in \mathbb{R}^3$ is the relative position to the target expressed in the body frame, $\mathbf{q} \in \mathbb{R}^4$ is the orientation as a unit quaternion, $\mathbf{v} \in \mathbb{R}^3$ is the linear velocity, $\boldsymbol{\omega} \in \mathbb{R}^3$ is the angular velocity, $\mathbf{a}_{\text{prev}} \in \mathbb{R}^{D_{\text{action}}}$ is the previous action taken, $v_{\text{des}} \in \mathbb{R}$ is the desired speed and $\Phi(\Delta t_{\text{lidar}}) \in \mathbb{R}^{D_{\text{temp}}}$ is the output of the Temporal Encoding Module.

The action space $\mathcal{A} \subset \mathbb{R}^{D_{\text{action}}}$ defines the high-level control commands for AAVs. The policy network outputs these commands, which are then tracked by a low-level controller.

\subsubsection{Reward Function}
\label{subsection:Reward_Function}

 The reward function $R_t = R(s_t, a_t)$ is composed of dense and sparse components, formulated as $R_t = \sum_i w_i r_i + r_T$, where the first term is a weighted sum of dense reward components $r_i$ with corresponding weights $w_i$, and $r_T$ is a sparse terminal reward that is non-zero only upon episode termination. The dense components are defined as follows:
\begin{enumerate}[label=\alph*), wide]
\item Static Safety Reward ($r_{\text{static}}$): This reward penalizes proximity to static obstacles based on LiDAR measurements. The calculation is defined by:
\begin{align}
    p_{\text{beam}} &=  \tanh\left(k \cdot (\frac{\min(d_{\text{obs}}, L_s)}{L_s}  - c)\right) + 1 , \label{eq:penalized_dist} \\
    r_{\text{static}} &= Q_{q}\left( \log(p_{\text{beam}}) \right) ,\label{eq:final_reward}
\end{align}
where $d_{obs}$ denotes the distance to obstacles, and $L_s$ is the threshold distance at which the penalty becomes active. 
The penalty curve, parameterized by $k$ and $c$, employs $\tanh$ and $\log$ functions to induce a sharp transition that strongly discourages collisions. 
The quantile function $Q_q$ ensures robustness to measurement noise.

\item Velocity Reward ($r_{\text{velocity}}$): This term encourages the robot to move towards the goal at a desired speed. It penalizes any misalignment between the velocity vector $\mathbf{v}$ and the unit direction vector $\hat{\mathbf{g}}$ towards the goal. It also manages the speed $v = \|\mathbf{v}\|$, encouraging the agent to maintain a speed near  $v_{\text{des}}$ within a desired range $[k_{v1}v_{\text{des}}, k_{v2}v_{\text{des}}]$, while penalizing speeds outside this range, where $0<k_{v1}<1<k_{v2}$.
The reward is formulated as:
    \begin{equation}
    \begin{split}
        r_{\text{velocity}} = &\mathbf{v} \cdot \hat{\mathbf{g}}  - (\max(0, v - k_{v2}v_{\text{des}}))^2 \\
        & - (\max(0, k_{v1}v_{\text{des}} - v))^2 \\
        & + \mathbb{I}(k_{v1}v_{\text{des}} \le v \le k_{v2}v_{\text{des}}) \\
        & \cdot \exp\left(-\frac{(v - v_{\text{des}})^2}{2\sigma^2}\right),
    \end{split}
    \end{equation}
    where $\mathbb{I}(\cdot)$ is the indicator function and $\sigma$ controls the reward shape.

    \item Height Penalty ($r_{\text{height}}$): A penalty is applied for altitude $z$ straying from a vertical corridor $[z_{\min}, z_{\max}]$, given by
    \begin{equation}
    \begin{split}
        r_{\text{height}} =  - (\max(0, z - z_{\max}))^2 
         - (\max(0, z_{\min} - z))^2.
    \end{split}
    \end{equation}
    
    \item Attitude Penalty ($r_{\text{attitude}}$): Excessive pitch $\alpha_{y}$ or roll $\alpha_{x}$ angles beyond a threshold $\alpha_{\max}$ are penalized through the term
    \begin{equation}
    \begin{split}
        r_{\text{attitude}} = & - (\max(0, |\alpha_{y}| - \alpha_{\max}))^2 \\
        & - (\max(0, |\alpha_{x}| - \alpha_{\max}))^2.
    \end{split}
    \end{equation}
\end{enumerate}

In addition to these dense rewards, large sparse rewards or penalties are issued at episode termination: a positive reward $r_{\text{goal}}$ for reaching the target, a negative penalty $r_{\text{collision}}$ for collision, and a penalty $r_{\text{limit}}$ for violating an operational boundary limit, such as exceeding maximum altitude or leaving the designated flight area.

\subsection{Asynchronous End-to-end Network with Temporal Encoding Module}
\label{subsection:Asynchronous end-to-end network}
Conventional end-to-end control suffers from the mismatch between high-rate IMU-based control and low-rate perception from sensors like LiDAR, whereas the proposed asynchronous architecture decouples them, enabling high-frequency control. This decoupling, however, introduces data staleness. This staleness is formally modeled via the Age of Information (AoI)~\cite{kaul2012real}---specifically, the age upon decision (AuD)---defined as 
\begin{equation}
\Delta t_{\text{lidar}} = t^i_{\text{ctrl}} - t_{\text{meas}},
\end{equation}
where $t^i_{\text{ctrl}}$ is the $i^{th}$ decision time and $t_{\text{meas}}$ is the measurement timestamp. 
The resulting high Age of Information (AoI) induces partial observability, breaking the Markov assumption in the original state space. To address this, a Temporal Encoding Module (TEM) that encodes the AoI is proposed as a policy input $\Phi(\Delta t_{\text{lidar}})$.

Let the standard observation 
\begin{equation}
    O'_t= [\mathbf{z}_t, \mathbf{p}_{\text{rel}}, \mathbf{q}, \mathbf{v}, \boldsymbol{\omega}, \mathbf{a}_{\text{prev}}, v_{\text{des}}]^T ,
\end{equation}
 which is the full state $s_t$ from Eq.\ref{eq:state_vector} without temporal information. The augmented observation $O_t$ is defined as the concatenation of $O'_t$ and the encoded delay feature:
\begin{equation}
    O_t := (O'_t, \Phi(\Delta t_{\text{lidar}})) = s_t.
    \label{eq:augmented_observation}
\end{equation}

By conditioning on its own velocity and the explicit time delay, the policy can learn to implicitly predict how the perceived environment has changed, compensating for the information staleness. Theoretically, augmenting the observation with the AoI reduces the conditional entropy of the state estimation ($H(S_t | O_t) \le H(S_t | O'_t)$) and, by the Law of Total Variance, eliminates the excess variance caused by delay uncertainty, thereby facilitating more stable policy learning.

\subsection{Network Architecture}
\label{subsection:Network_Architecture}

Since the LiDAR perception data is structured as a 2D pseudo-image, a convolutional neural network (CNN) is utilized as a backbone to extract spatial features and transform them into a compact feature embedding. This perception embedding is then concatenated with the robot's internal state vector. This combined state is then processed by a MLP decoder to generate a unified feature representation for the subsequent actor and critic networks.

From this unified representation, the actor network outputs $(\alpha, \beta)$ for a Beta distribution to generate normalized continuous actions, which improves convergence in constrained spaces~\cite{chou2017improving}. A constant offset $\epsilon$ is added after \texttt{Softplus} to ensure $\alpha, \beta > \epsilon$, maintaining a unimodal distribution and avoiding unstable U-shaped behaviors.

\subsection{Policy Training}
\label{subsection:Policy_Training}
A primary challenge in this asynchronous framework is the temporal discrepancy between low-frequency perception updates and high-frequency control actions. This mismatch provides a weak supervisory signal for the policy, making direct training inefficient. To overcome this challenge, a two-stage curriculum learning strategy is adopted.

\textbf{I. }Synchronous Training Stage: The network is trained synchronously within the simulator using ideal, high-frequency perception data. Within this idealized setting, the policy acquires a fundamental baseline navigation capability. Notably, throughout this learning process, the simulated AoI is maintained to be zero.
    
\textbf{II. }
Asynchronous Training Stage: After convergence in the synchronous phase, training shifts to an asynchronous paradigm where the simulator provides low-frequency perception data to mimic real-world constraints. Unlike the synchronous stage (AoI = 0), the simulated AoI is now a non-zero, time-varying value that quantifies data staleness. The curriculum-based training strategy promotes robust policy generalization by employing the Temporal Encoding Module (TEM) to adapt decision-making under time-varying perception delays. This mechanism effectively compensates for the performance degradation induced by transmission latency.

Crucially, the synchronous training in the first stage provides the policy with a critical warm start, making the task of learning to leverage the TEM to handle time-varying delays in the second stage more tractable and thereby significantly improving training stability.

\begin{table*}[!h]
\centering
\caption{PERFORMANCE BENCHMARK UNDER VARYING SENSOR FREQUENCIES. All trials are conducted in an environment with an obstacle density of $0.2~{m}^{-2}$ and a target average velocity of $2.5~m/s$.}

\label{tab:champion_models_ranking}
\setlength{\tabcolsep}{6pt}
\begin{tabular}{@{}l c c c c c c c@{}}
\toprule
\multirow{2}{*}{\textbf{Model}} & \multicolumn{2}{c}{\textbf{Reach Goal ($\uparrow$)}} & \multirow{2}{*}{\textbf{Performance Loss ($\downarrow$)}} & \multicolumn{2}{c}{\textbf{Avg. Velocity ($m/s$) }} & \multicolumn{2}{c}{\textbf{Onboard Processing Demand}} \\
\cmidrule(lr){2-3} \cmidrule(lr){5-6} \cmidrule(lr){7-8}
 &   High Freq &   Low Freq & &   High Freq &   Low Freq &   High Freq &   Low Freq \\
\midrule
NavRL~\cite{xu2025navrl} & 86.96\% & 75.37\% & 11.6\% & 2.49 & 2.46 & High & Low \\
YOPO~\cite{lu2024you} & 67.71\% & 58.51\% & 9.2\% & 2.51 & 2.52 & Medium & Low \\
EGO-Planner-v2~\cite{zhou2022swarm} & 79.87\% & 76.71\% & 3.2\% & 2.53 & 2.53 & Medium & Medium \\
Ours (Sync. Baseline / Async) & \textbf{93.67\%} & \textbf{91.08\%} & \textbf{2.6\%} & {2.54} & {2.54} & Medium & Low \\
\bottomrule
\end{tabular}
\end{table*}

\section{RESULTS AND DISCUSSIONS}

\subsection{Implementation Details}
\label{subsection:Implementation_Details}

\subsubsection{Network Architecture and Training}

\begin{figure}[t] 
    \centering
    \includegraphics[width=0.88\linewidth]{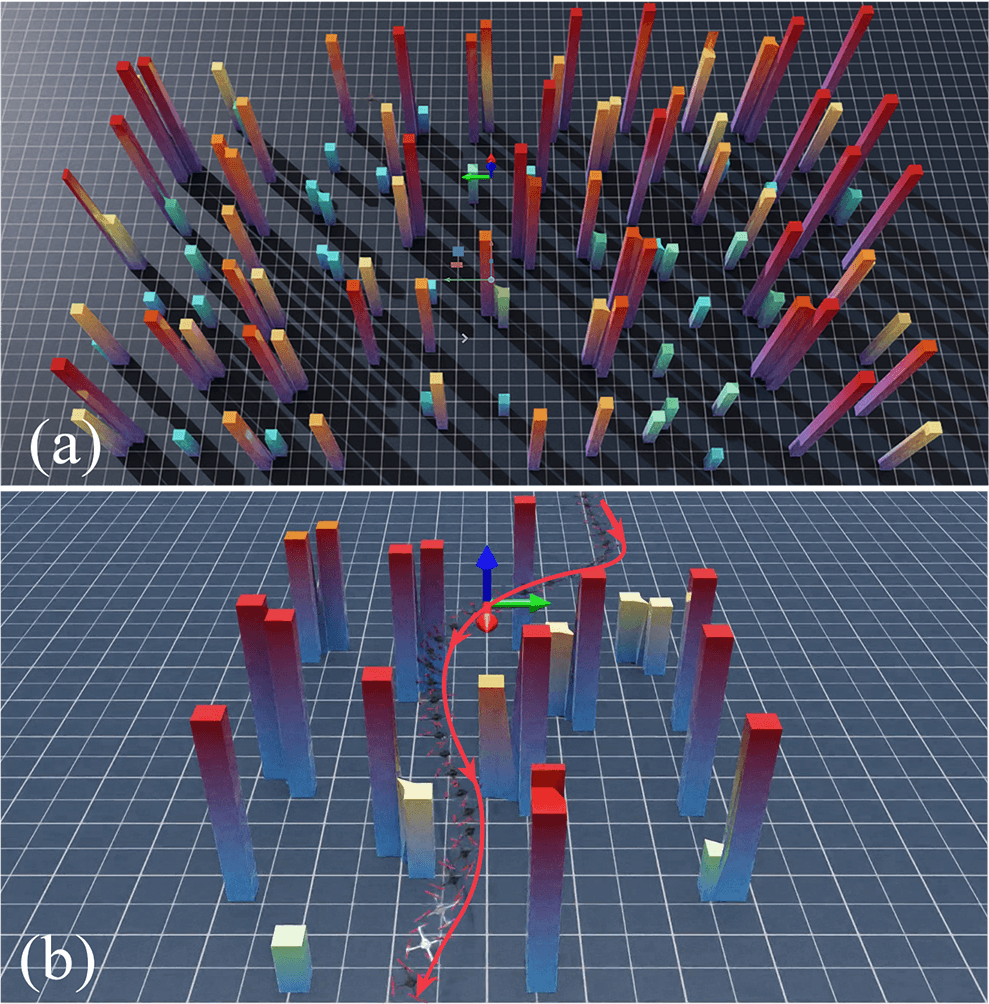} 
    
    \caption{Visualization of (a) a sample training environment and (b) the collision-free trajectory generated by the proposed end-to-end navigation model.}
    \label{fig:training_setup}
    \vspace{-3mm} 
\end{figure}

In this paper, the actor-critic policy is optimized using the Proximal Policy Optimization (PPO) algorithm~\cite{schulman2017proximal}. Training is conducted in a forest-like simulation environment developed with the Omnidrones framework~\cite{xu2024omnidrones} in NVIDIA Isaac Sim. The policy network is trained on a single NVIDIA RTX 5880 Ada GPU using 4000 parallel simulation environments as shown in Fig.~\ref{fig:training_setup}, with a desired speed randomly sampled from $[1, 4]~{m/s}$. The network's perception input is the LiDAR point cloud, which covers an azimuthal range of $[-\frac{1}{2}\pi, \frac{1}{2}\pi)$
with resolutions of $\Delta\theta_{\text{pillar}}=\frac{1}{60}\pi$ and $\Delta\phi_{\text{pillar}}=\frac{1}{36}\pi$.
The convolutional backbone processes this pseudo-image and outputs a compact feature embedding $z_{t} \in \mathbb{R}^{32}$ 
.
The policy network's action space is defined as target linear velocities in the robot's body frame.
The non-perceptual state input is composed of the action vector from the previous timestep, 
other proprioceptive states, temporal feature from the Temporal Encoding Module (TEM), which processes the AoI and is implemented as a sinusoidal encoder inspired by~\cite{vaswani2017attention}.

For a continuous AoI $\Delta t_{\text{lidar}}$, the feature vector $\Phi(\Delta t_{\text{lidar}})$ is computed as:
\begin{equation}
\label{eq:tem}
\Phi(\Delta t_{\text{lidar}}) = [\sin(t_j), \cos(t_j), \sin(t_j/100), \cos(t_j/100)]^T,
\end{equation}
where the discrete step $j$ and re-quantized time $t_j$ are derived using $j = \text{round}(\Delta t_{\text{lidar}}/\Delta t)$ with a resolution of $\Delta t = 0.01\text{s}$, and $t_j = j \cdot \Delta t$.
The policy is trained using the AdamW optimizer and a PPO clip ratio of 0.1. 

\subsubsection{Real-World Experimental Platform}
For real-world validation, the trained policy is deployed on a custom-built quadrotor, as shown in Fig.~\ref{fig:first}(a). Onboard computation is handled by an Intel NUC 13.
This unit interfaces with an NxtPx4 flight controller. Environmental perception is provided by a Livox Mid-360 LiDAR~\cite{livox:mid360:specs}. 
An onboard Intel RealSense D435i camera is used solely for visualization and does not contribute to the navigation algorithm.
To evaluate the framework's performance and portability on different embedded systems, the same policy is also successfully deployed and tested on an NVIDIA Jetson Orin NX module.

\subsection{Simulation Experiments}

\subsubsection{Benchmark with Previous Systems}

A benchmark analysis is conducted to evaluate the proposed framework (a representative successful trajectory is shown in Fig.~\ref{fig:training_setup}(b)) against state-of-the-art methods: two Learning-based (NavRL~\cite{xu2025navrl}, YOPO~\cite{lu2024you}) and one optimization-based (EGO-Planner-v2~\cite{zhou2022swarm}) with each method evaluated 2000 trials. For each trial, the environment is randomly initialized, featuring a high-density obstacle field ($0.2~{m}^{-2}$) composed of rectangular prisms with horizontal side lengths sampled from $[0.4, 0.6]~m$. The agent's task is to navigate a path at a target average velocity of $2.5~{m/s}$. To assess robustness to sensor constraints, we evaluated performance under both an ideal $100~Hz$ and a constrained $10~Hz$ perception rate, with the latter chosen to mirror the native update rate of the LiDAR sensor used in our physical tests~\cite{livox:mid360:specs}. To ensure a fair comparison, the underlying low-level controller for all tested methods is maintained at a constant $100~Hz$, and, within each perception-rate regime, all methods share the same compute budget.

\begin{figure*}[!ht]
    \centering
    \includegraphics[width=0.94\linewidth]{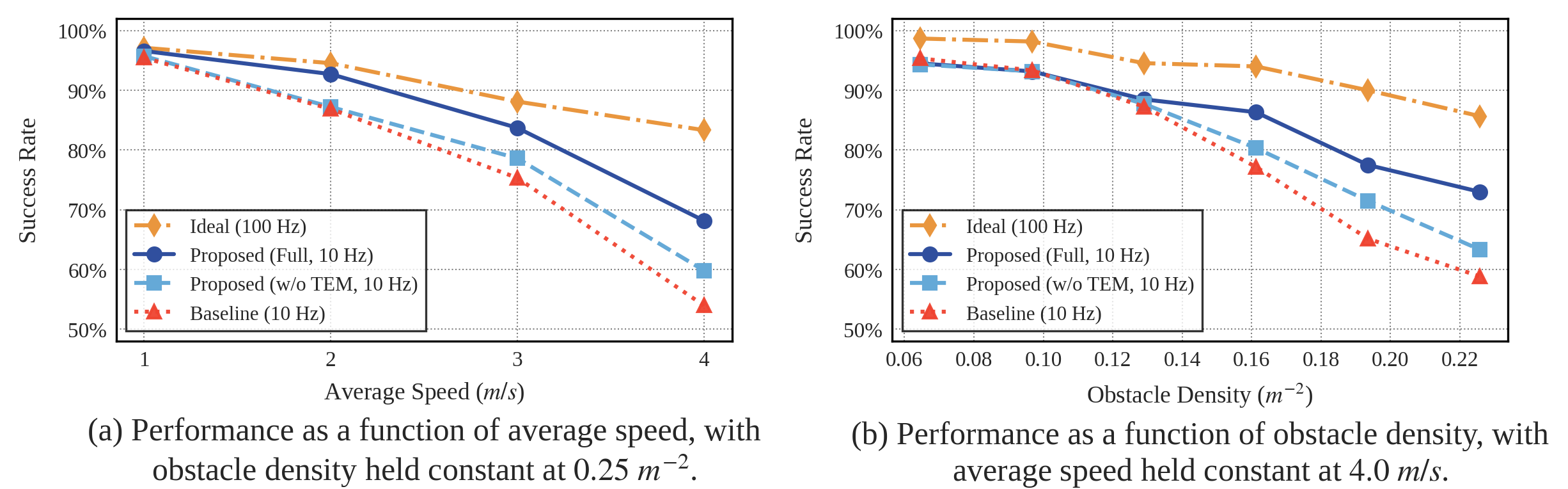}
    \caption{Evaluation of model success rate under challenging conditions.}
    \label{fig:obstacles_plot}
    \vspace{-3mm} 
\end{figure*}

The results in Table~\ref{tab:champion_models_ranking} highlight our framework's exceptional robustness to varying sensor frequencies. At a $100~Hz$ perception rate, our method achieves a success rate of 93.67\%, which remained remarkably high at 91.08\% (a minimal 2.6\% drop) when the rate is reduced to $10~Hz$. This stability stands in stark contrast to competing synchronous systems, such as NavRL~\cite{xu2025navrl}, which suffers a much larger 11.6\% performance degradation. This performance gap arises because the control frequency of conventional learning-based approaches is fundamentally coupled to the perception rate. Similarly, optimization-based planners like EGO-Planner-v2~\cite{zhou2022swarm}, despite featuring a decoupled core computation, are ultimately bottlenecked by real-time map generation from the same slow data stream. In contrast, our framework's robustness is a direct consequence of its asynchronous architecture. By decoupling the perception and control loops, our design enables high-frequency policy execution while using the Temporal Encoding Module (TEM) to actively compensate for information delays, thereby overcoming the bottlenecks inherent to synchronous models.

\subsubsection{Ablation Studies and Performance Analysis}

To validate the proposed asynchronous framework, two sets of simulation experiments are conducted to analyze performance under varying speeds and obstacle densities, with 1000 trials for each tested condition. The environment consists of a 20-meter path populated with obstacles shaped as rectangular prisms, whose horizontal side lengths are sampled from 0.4 m to 0.6 m. The evaluation compares the following models:
\begin{itemize}
    \item \textbf{Proposed Model}: The complete asynchronous framework with the sinusoidal TEM.
    \item \textbf{Ideal Case}: A synchronous version of the proposed model with an ideal high-frequency ($100~Hz$) perception input, serving as a performance upper bound.
    \item \textbf{Ablation Model}: The proposed model with the TEM removed, to specifically isolate its contribution.
    \item \textbf{Baseline Model}: A conventional synchronous end-to-end model, where the control loop frequency is limited by the LiDAR sensor ($10~Hz$).
\end{itemize}
In the first experiment (Fig.~\ref{fig:obstacles_plot}(a)), performance is evaluated as a function of average speed, with obstacle density held constant at $0.25~{m}^{-2}$. In the second (Fig.~\ref{fig:obstacles_plot}(b)), robustness is analyzed by varying the obstacle density while maintaining a constant high speed of $4~m/s$.

The findings indicate that conventional synchronous models, though effective in simple settings, degrade in high-speed, high-density scenarios, whereas our asynchronous method surpasses the baseline by over 14 percentage points. This performance is further enhanced by the Temporal Encoding Module (TEM); its removal caused a significant ($p < 0.05$) performance drop of 8.4 and 9.7 percentage points under the maximum speed and maximum density conditions, respectively. Notably, the TEM's contribution becomes more pronounced as the task grows more challenging, with the performance gap widening significantly at higher speeds and densities. Crucially, the complete framework also closes a significant portion of the performance gap to an ideal, fully synchronous system, validating it as a robust and effective solution for high-frequency, reactive navigation.

\begin{figure*}[t] 
    \centering
    \includegraphics[width=0.91\linewidth]{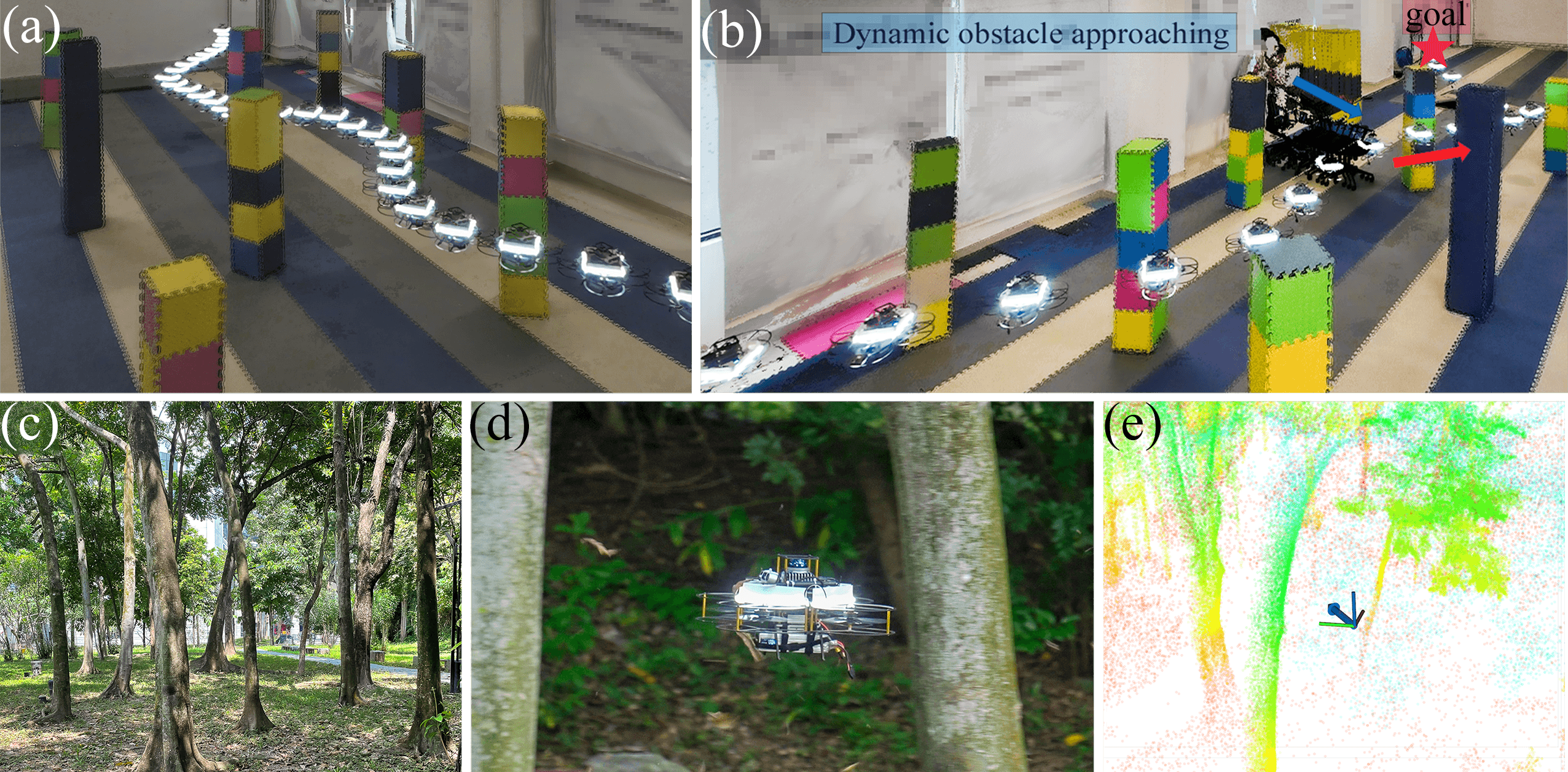} 
    \caption{Real-world flight validation of our asynchronous framework, demonstrating successful zero-shot sim-to-real transfer in cluttered environments. (a) The AAV navigating a dense indoor obstacle field ($0.25~{m}^{-2}$). (b) The AAV's flight trajectory in a scenario involving a dynamic obstacle. (c) First-person and (d) third-person views of autonomous navigation through a dense forest ($0.18~{m}^{-2}$). (e) Onboard visualization displaying the input LiDAR point cloud alongside the corresponding velocity command (blue arrow) generated by the policy.}
    \label{fig:real}
    \vspace{-3mm} 
\end{figure*}

\subsection{Physical Flight Tests and Performance Analysis}

The robustness of our framework is validated by its successful zero-shot sim-to-real transfer, deploying a policy trained entirely in simulation directly onto the physical platform without any fine-tuning. For these tests, the AAV relies solely on a low-frequency LiDAR operating at $10~Hz$ (the typical native update rate of the onboard Livox Mid-360 sensor~\cite{livox:mid360:specs}) while navigating two distinct and challenging environments at an average speed of $1.3~m/s$, with a maximum velocity of $2.0~m/s$. Multiple successful trials are conducted in each environment to confirm the policy's consistency. During these flights, the policy consistently operates with a high Perception AoI, frequently exceeding $100~ms$, which validates the effectiveness of the TEM in a real-world setting.

In a cluttered indoor space ($0.25~{m}^{-2}$ density), the AAV successfully navigates the planned course and demonstrates reactive avoidance against dynamic objects, with key moments documented in Fig.~\ref{fig:real}(a-b). Similarly, in a dense outdoor forest ($0.18~{m}^{-2}$ density), the drone consistently and autonomously maneuvers through the trees. Its flight and onboard decision-making process are visualized from multiple perspectives in Fig.~\ref{fig:real}(c-e), and a complete representative trajectory is presented in Fig.~\ref{fig:first}(b).

\begin{table}[!tb]
    \centering
    \begin{threeparttable}
        \caption{COMPUTATIONAL LATENCY ON DESKTOP AND ONBOARD PLATFORMS}
        \label{tab:performance_final_compact}
        \small
        \renewcommand{\arraystretch}{1.2}
        \setlength{\tabcolsep}{5pt}
        
        \begin{tabular}{@{} >{\RaggedRight}p{2.8cm} rrrr @{}}
            \toprule
            & \multicolumn{2}{c}{\textbf{Desktop}\tnote{*}} & \multicolumn{2}{c}{\textbf{Onboard}\tnote{\dag}} \\
            \cmidrule(lr){2-3} \cmidrule(l){4-5}
            \textbf{Module} & \textbf{GPU} & \textbf{CPU} & \textbf{GPU} & \textbf{CPU} \\
            \midrule
            \multicolumn{5}{l}{\textit{\textbf{Base Module Cost (Time in ms)}}\tnote{\S}} \\
            \textbf{Perception Total} & \textbf{0.40}\tnote{\ddag} & \textbf{0.80}\tnote{\ddag} & \textbf{3.99}\tnote{\ddag} & \textbf{1.15}\tnote{\ddag} \\
            \quad - PC Pre-proc. & 0.26 & 0.54 & 2.51 & 0.80 \\
            \quad - Perc. Network & 0.14 & 0.26 & 1.48 & 0.35 \\
            \textbf{Control Policy} & \textbf{0.39} & \textbf{0.21} & \textbf{1.72} & \textbf{0.27} \\
            \bottomrule
        \end{tabular}
        
        \begin{tablenotes}
            \item[*] \footnotesize Desktop: Intel Core i5-14600K CPU, NVIDIA RTX 4060 Ti GPU.
            \item[\dag] \footnotesize Onboard: Intel NUC 13 i7-1360P CPU, NVIDIA Jetson Orin NX GPU.
            \item[\S] \footnotesize Timings reflect pure computational latency, excluding overhead from data I/O and transmission.
            \item[\ddag] \footnotesize Totals assume sequential execution.
        \end{tablenotes}
    \end{threeparttable}
\end{table}

The computational performance that underpins this reliable flight behavior is detailed in Table~\ref{tab:performance_final_compact}. The analysis shows the asynchronous architecture leverages the high efficiency of the Control Policy network by executing it multiple times for each slower perception update. This design proves effective in practice, sustaining a stable $100~Hz$ control loop during physical deployment on both an Intel NUC 13 and an NVIDIA Jetson Orin NX.

The physical flight tests demonstrate our framework's viability for autonomous navigation on resource-constrained AAVs. The policy's robustness is evidenced by its successful zero-shot transfer to complex, real-world environments. This success stems from three key factors. First, the inherent robustness of our perception front-end stems from its high degree of environmental abstraction. Second, a lightweight network architecture enables real-time onboard inference. Third, the asynchronous design preserves a high-frequency control loop even under low-rate perception. These results indicate that our approach is a practical and effective solution for deployment in real-world robotic systems.

\subsection{Limitations and Future Work}
\label{subsection:limitations}
While the proposed framework demonstrates robust performance, two key limitations exist. 
First, as a purely reactive method, the current policy lacks an explicit trajectory prediction module for dynamic agents. Consequently, while it handles static or slowly moving obstacles effectively, reliably avoiding high-speed dynamic obstacles remains a challenge.
Second, regarding the sim-to-real gap at high speeds, the discrepancy between the simulated and physical platform dynamics becomes pronounced as velocity increases. Achieving robust flight at higher velocities requires precise system identification to align the low-level controller's response, which is a critical direction for future work.

\section{CONCLUSIONS}

This paper introduces an end-to-end framework that successfully mitigates the temporal conflict between low-frequency perception and high-frequency control in AAV navigation. The success of our approach is built upon three key innovations: a lightweight end-to-end architecture with a computationally efficient LiDAR processing module; a principled asynchronous design featuring a Temporal Encoding Module to manage the Age of Information; and a two-stage curriculum that ensures effective policy training. The framework's robustness is validated through successful zero-shot sim-to-real deployment, where the policy sustains a $100~Hz$ control rate on an onboard computer while navigating a high-density forest. By enabling high-frequency control with low-frequency sensors, this work directly addresses key limitations of end-to-end models, significantly enhancing their agility and robustness for deployment in cluttered, real-world environments.

\addtolength{\textheight}{-12cm}   





\bibliographystyle{IEEEtran}
\bibliography{references}{}

\end{document}